\documentclass[letterpaper, 10pt, conference]{ieeeconf}  

\IEEEoverridecommandlockouts                              

\usepackage{amsthm}
\usepackage{times}
\usepackage{multicol}
\usepackage[bookmarks=true]{hyperref}
\usepackage{xcolor}
\usepackage{hyperref}
\usepackage{amsmath, amssymb}
\usepackage{amsfonts}
\usepackage{graphicx}
\usepackage{siunitx}
\usepackage{standalone}
\usepackage{booktabs}
\usepackage[ruled,vlined,linesnumbered,noend]{algorithm2e}
\usepackage{mdframed}
\usepackage{fancyvrb,multirow}
\usepackage{soul}
\usepackage{dsfont,mathabx}
\usepackage{array, booktabs}
\usepackage{booktabs}
\usepackage{makecell}
\usepackage{threeparttable}
\usepackage{float}
\usepackage{stfloats}
\usepackage{mathrsfs}
\usepackage{lipsum}
\usepackage{cleveref}

\theoremstyle{definition}

\title{\LARGE \bf
LOMORO: Long-term Monitoring of Dynamic Targets with Minimum Robotic Fleet
under Resource Constraints}

\author{Mingke Lu, Shuaikang Wang and Meng Guo
  \thanks{The authors are with the College of Engineering,
    Peking University, Beijing 100871, China.
    This work was supported by the National Natural Science Foundation
    of China (NSFC) under grants 62203017, U2241214, T2121002.
    Contact: {\tt\small meng.guo@pku.edu.cn}.}
}

\begin{document}
\maketitle
\thispagestyle{empty}
\pagestyle{empty}


\begin{abstract}
  Long-term monitoring of numerous dynamic targets can be tedious for a human operator
  and infeasible for a single robot,
  e.g., to monitor wild flocks, detect intruders, search and rescue.
  Fleets of autonomous robots can be effective by acting collaboratively and concurrently.
  However, the online coordination is challenging due to the unknown behaviors
  of the targets and the limited perception of each robot.
  Existing work often deploys all robots available without minimizing the fleet size,
  or neglects the constraints on their resources such as battery and memory.
  This work proposes an online coordination scheme called LOMORO
  for collaborative target monitoring, path routing and resource charging. 
  It includes three core components:
  (I) the modeling of multi-robot task assignment problem under
  the constraints on resources and monitoring intervals;
  (II) the resource-aware task coordination algorithm
  iterates between the high-level assignment of dynamic targets  
  and the low-level multi-objective routing via the Martin's algorithm;
  (III) the online adaptation algorithm in case of unpredictable target behaviors
  and robot failures.
  It ensures the explicitly upper-bounded monitoring intervals for all targets
  and the lower-bounded resource levels for all robots,
  while minimizing the average number of active robots.
  The proposed methods are validated extensively via large-scale simulations
  against several baselines,
  under different road networks, robot velocities, charging rates and monitoring intervals. 
\end{abstract}
\section{Introduction}\label{sec:intro}
Mobile robots such as unmanned aerial vehicles (UAVs)
and unmanned ground vehicles (UGVs) are becoming more
capable of autonomous inspection and navigation.
Via wireless communication and collaboration,
a fleet of such robots can be deployed to monitor
large areas that are otherwise too demanding for human operators,
e.g., to explore unknown territory~\cite{burgard2005coordinated},
and track moving targets~\cite{tzes2023graph,kalluraya2023multi}.
Particularly, active monitoring of unknown dynamic targets has attracted significant attention,
see~\cite{atanasov2014information,dames2012decentralized,
  chung2006decentralized,schlotfeldt2018anytime}.
It incorporates several challenging aspects of
multi-robot coordination:
(I) the behavior of each target regarding its velocity and future path
is uncertain, meaning that the assignment of targets to robots should be adaptive;
(II) to ensure the accuracy of monitoring, a minimum
monitoring interval for each target is required.
Moreover, for long-term (possibly indefinite) mission,
it is inevitable that the robots are subject to resource constraints
such as battery and memory.
In this case, the planning of monitoring tasks
and charging activities are heavily dependent
and should be planned as a whole.
Existing methods~\cite{tzes2023graph, kalluraya2023multi,
  atanasov2014information, kantaros2019asymptotically, le2009trajectory,
  sung2020distributed, tokekar2014multi, zhou2019sensor}
mostly consider the short-term monitoring task of
one or several targets with known behaviors,
via a fixed fleet size of robots without resource constraints.
It remains an open problem to coordinate online the minimum fleet
of robots, under the strict constraints of minimum monitoring intervals
over targets and the minimum resources over robots.

\begin{figure}[t!]
  \centering
  \includegraphics[width=0.98\linewidth]{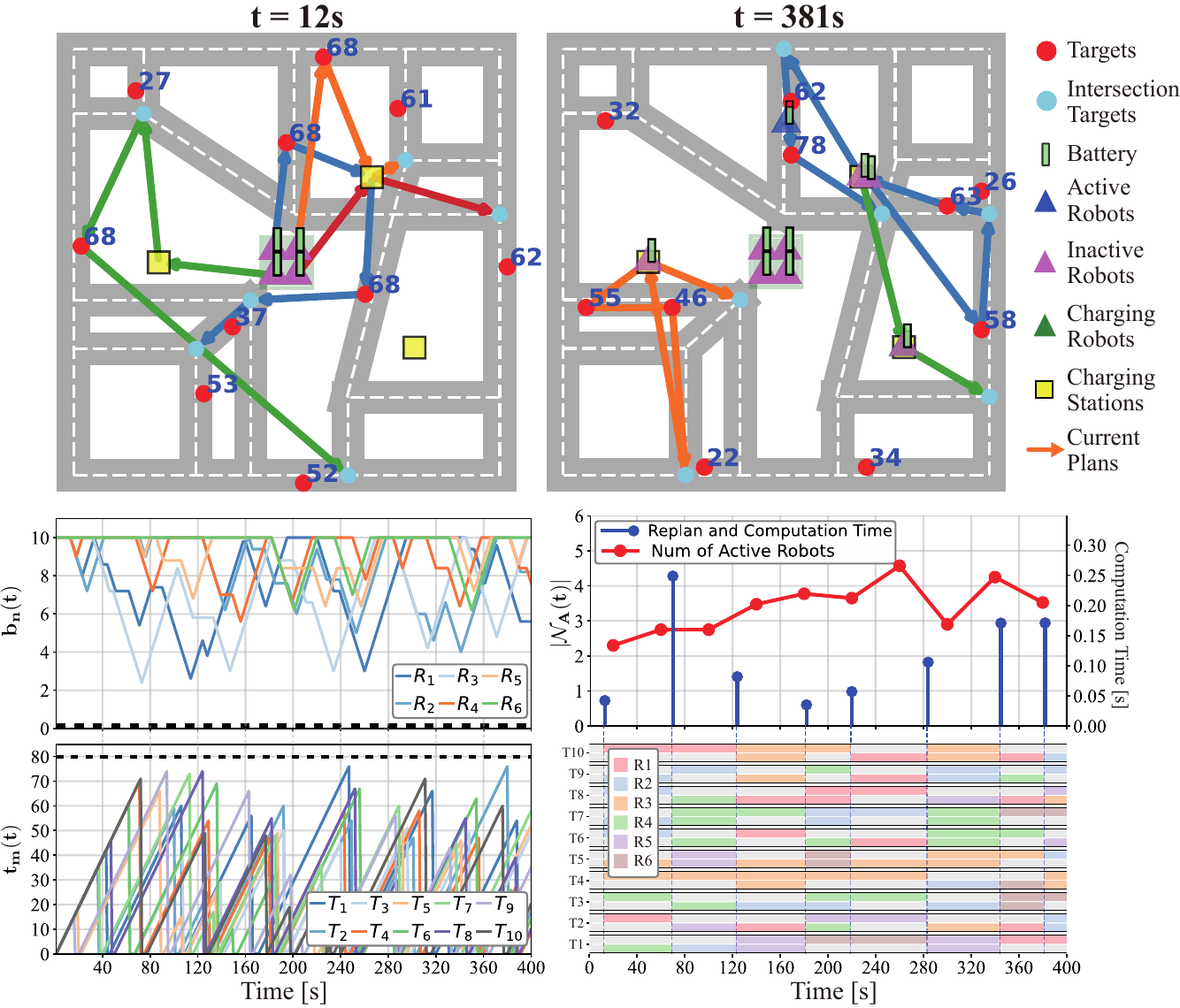}
  \vspace{-3mm}
  \caption{Illustration of the considered scenario.
    \textbf{Top-Left} and \textbf{Top-Right}: $3-4$ UAVs are actively monitoring $10$ targets (in red) within a road network, with their static (t=$\SI{12}{s}$) and online (t=$\SI{381}{s}$) plans.
    \textbf{Middle Left}: Batteries of 6 robots during the online execution of $\SI{400}{s}$.
    \textbf{Bottom-Left}: Intervals from the last monitoring of 10 targets.
    \textbf{Middle-Right}: Average number of active robots, the time when replans take place, and their computation time.
    \textbf{Bottom-Right}: The robots responsible for each target during any consecutive replans. Each target has 2 rows, with the lower representing the target node and the upper the intersection target node respectively.}

  \label{fig:overall}
  \vspace{-5mm}
\end{figure}

\subsection{Related Work}\label{subsec:intro-related}
\subsubsection{Active Information Acquisition}
Active information acquisition for sensing robots, originating
in \cite{atanasov2014information, le2009trajectory}, focuses on optimizing
robot motion to maximize information gain. Early approaches employ offline
search algorithms that explore the joint state-information space through
forward value iteration and its reduced variant, leveraging separation principles
under linear observation models. These principles have been extended to
multi-robot systems: centralized, non-myopic solutions using sampling-based
methods are proposed in \cite{kalluraya2023multi, kantaros2019asymptotically},
while decentralized, myopic strategies are investigated
in \cite{dames2012decentralized,chung2006decentralized,schlotfeldt2018anytime,
  chen2024accelerated}. Learning-based techniques, such as those in
\cite{tzes2023graph}, train distributed policies to mimic optimal planners.
However, existing methods predominantly assume fixed robot team sizes and
targets with known dynamics or control inputs. A key unresolved challenge lies
in dynamically adjusting the robot team size to monitor targets exhibiting
unknown behaviors, including velocities and road network navigation paths.

\subsubsection{Multi-robot Task Assignment}
In contrast, multi-target monitoring has also been framed as a sequential
high-level assignment problem, with robots assigned to targets to minimize uncertainty.
While \cite{chopra2017distributed} introduces a distributed Hungarian method,
its scope is restricted to one-to-one assignments.
Several works, including \cite{sung2020distributed, tokekar2014multi, zhou2019sensor},
rephrase this as a simultaneous action and target assignment problem,
proposing distributed approximation algorithms such as linear programs.
These methods, however, typically assume known target trajectories and
rely on synchronized robot motions with predefined primitives.
The complementary work of \cite{dames2017detecting} maximizes the number of
tracked targets using a fixed robot team via a 2-approximation greedy approach.
Recent studies like \cite{zhou2018resilient, zhou2023robust} investigate
robust assignments against communication or sensing attacks but retain
assumptions of synchronized motions with finite primitives. Departing from
these frameworks, this work determines online the flexible assignment of
robots to targets according to actual observations that can not be
determined offline.

\subsubsection{Planning under Resource Constraints}
Recent advances in robot motion and task planning under resource constraints
emphasize optimization-based frameworks that balance computational efficiency
with energy and memory limitations. State-of-the-art approaches integrate
multi-objective optimization to handle battery constraints, often leveraging
dynamic programming~\cite{di2015energy} or model predictive
control~\cite{seisa2022edge} to allocate energy budgets while ensuring task
completion. Memory-aware planning algorithms~\cite{guo2018resource}
or hierarchical task decomposition~\cite{homberger2007multi},
reduce computational overhead
by pruning redundant states or compressing environment representations.
For distributed systems, decentralized strategies in~\cite{afrin2021resource} use
reinforcement learning to optimize local decisions under shared resource
limits. Hybrid methods combining offline pre-planning with online
adaptation, such as the anytime algorithms in \cite{mourikis2006optimal},
dynamically adjust plans based on real-time resource consumption.
Despite progress, challenges persist in scaling these methods to highly dynamic
large-scale environments and flexible fleet size.

\subsection{Our Method}\label{subsec:intro-our}

This work addresses the \emph{long-term} monitoring task of numerous dynamic
targets within a road network via a fleet of aerial robots,
where each target has a strict (potentially different) monitoring interval.
More importantly, the robots have limited resources
(such as battery and memory)
that are consumed over time and should be recharged often.
The targets follow a constant-velocity model,
but with unknown velocity on each road, and unknown path within the road network.
The proposed method consists of three components:
(I) an search-based assignment algorithm is designed that searches through
a partial sequence of robots, in which the subset of targets
are assigned in an incremental way.
It ranks the feasible assignments based on a multi-objective measure;
(II) a maximum-allowed Martin's algorithm (MAM) is proposed
to determine \emph{simultaneously} the optimal
subset and sequence of targets to monitor,
and the optimal charging station to charge,
via an efficient incremental label-setting procedure.
(III) an online adaptation scheme is proposed to monitor the feasibility
of the local plan of each vehicle online.
In case of violation of the above constraints,
it triggers the first component to find first an alternative assignment
that is feasible (by recruiting additional robots if needed),
then improves the quality as more planing time is permitted.
It is proven that the constraints of resources and monitoring intervals
are fulfilled at all time,
while the average number of robots that actively monitors the targets
is minimized.
Extensions such as free membership of targets and charging stations
that are dynamically moving are also demonstrated.
Extensive simulations are performed over large-scale fleets
and targets over complex scenes.

Main contribution of this work is three-fold:
(I) the novel formulation of the minimum-fleet monitoring problem
of unknown dynamic targets within road networks,
under strict constraints of monitoring interval and resources;
(II) the hierarchical solution that adapts the fleet size
and vehicle trajectories online,
according to real-time observations of the target behavior;
(III) the scalable solution that allows the deployment of a few UAVs
to monitor a large number of dynamic targets.
\begin{figure*}[t]
    \centering
    \setlength{\abovecaptionskip}{0.cm}
    \includegraphics[width=\linewidth]{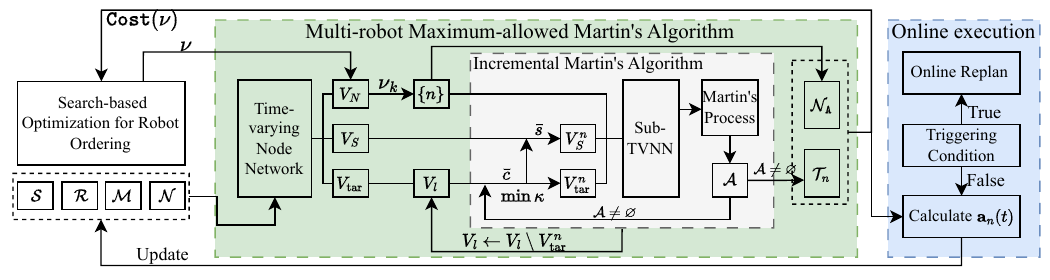}
    \vspace{-0.6cm}
    \caption{Illustration of the proposed framework, which consists of three parts: Search-based optimization for robot ordering, multi-robot maximum-allowed Martin's algorithm (MAM) and online execution. The MAM consists of the time-varying node network (TVNN) and the incremental Martin's algorithm (IMA). The IMA mainly consists of incremental sub-TVNN and the Martin's Process.}
    \label{fig:framework}
    \vspace{-0.6cm}
\end{figure*}
\section{Problem Description}\label{sec:problem}

\subsection{System Description}
Consider a workspace $\mathcal{W} \subset \mathbb{R}^2$ 
within which the robots and targets coexist. 
We define the environment as $\mathcal{P}\triangleq (\mathcal{R},\mathcal{S})$, 
where $\mathcal{R} \subset \mathcal{W}$ denotes the road network 
and $\mathcal{S}=\{1, 2,\cdots,S\}$ denotes the set of charging stations. 
The road network is defined as $\mathcal{R}\triangleq (\mathcal{V}, \mathcal{E})$, 
where $\mathcal{V} \subset \mathcal{W}$ is the set of intersections
and $\mathcal{E} \subset \mathcal{V} \times \mathcal{V}$ is a set of 
straight roads connecting these hubs. For each charging station $s \in \mathcal{S}$, 
its position is $\mathbf{z}_s\in \mathcal{W}$ and the charging capacity per unit time is $\beta_s>0$.
Moreover, there is a fleet of robots $\mathcal{N} \triangleq \{1,\cdots ,N\}$,
which is divided into two sets,
i.e., $\mathcal{N} \triangleq  \mathcal{N}_{\texttt{A}}(t) \cup \mathcal{N}_{\texttt{I}}(t),$
where $\mathcal{N}_{\texttt{A}}(t)$ denotes the 
active robots and $\mathcal{N}_{\texttt{I}}(t)$ denotes the inactive robots. 
The state of each robot~$n \in \mathcal{N}$ is given
by its position~$\mathbf{x}_n(t)\in \mathcal{W}$, 
velocity~$\mathbf{v}_n(t)\in \mathbb{R}^2$ and battery~$b_n(t)>0$.
Its rate of battery consumption is defined as 
$\gamma_n(t)\triangleq \gamma(||\mathbf{v}_n(t)||), t\geq0$,
where~$\gamma:\mathbb{R}\to \mathbb{R}$ is a monotonically increasing function,
and~$||\cdot||$ denotes the $L_2$ norm.
The maximum velocity is~$v_n^\text{max}>0$ and the battery capacity is $b_n^\text{max}>0$. 
We define the 
control space of the robots as $\mathcal{U}$, 
and the robot follows the nonlinear dynamics:
\begin{equation}    \label{eq:dynamics}
    [\mathbf{x}_n(t+1),\mathbf{v}_n(t+1)]^T \triangleq f(\mathbf{x}_n(t), \mathbf{v}_n(t),\mathbf{u}_n(t)), 
\end{equation}
where $f(\cdot)$ is the dynamic model and $\mathbf{u}_n(t) \in \mathcal{U}$ is the
control input of the robot $n$ at time $t \geq 0$.
In addition, the sensor range of robot~$n$ is $R_n>0$, within which it can
make observations and detect targets.

Given the set of targets~$\mathcal{M} \triangleq \{1,\cdots ,M\}$,
each target~$m\in \mathcal{M}$ moves on roads in $\mathcal{R}$, 
following the constant-velocity model~\cite{atanasov2014information}
but with different velocities, i.e.,
$\mathbf{y}_m(t+1) \triangleq \mathbf{y}_m(t)
+ h\cdot \mathbf{v}_m$,
where~$h > 0$ is the sampling time, $\mathbf{y}_m(t) \in 
\mathcal{W}$ is the target state at~$t \geq 0$, and $\mathbf{v}_m \in \mathbb{R}^2$ 
is the constant velocity of target~$m$.
We assume that both the number of targets $M$ and their initial states are known.
Namely, the target may randomly switch to a 
different road at each hub point~$e\in \mathcal{E}$, i.e., 
following an unknown path.

\subsection{Local Plans and Constraints}\label{subsec:constraint}
Each robot~$n \in \mathcal{N}_{\texttt{A}}(t)$ has a local plan~$\mathcal{T}_n$, 
which is a sequence of targets and charging stations assigned to~$n$. 
For each robot $n \in \mathcal{N}_{\texttt{A}}(t)$, we define $\mathbf{a}_n(t)$ 
as its action sequence, where 
$\mathbf{a}_n(t) \in \{a_{\text{tr}}(v), a_{\text{mo}}(m), a_{\text{ch}}(s)\},
 v\in\mathcal{M}\cup\mathcal{S}, m\in\mathcal{M},s\in\mathcal{S}, v,m,s \in \mathcal{T}_n$,
and $T_d(\cdot)$ is the duration of the action.
If $\mathbf{a}_n(t)=a_{\text{tr}}(v)$, robot~$n$ is during navigation to a target
or at a charging station.
If $\mathbf{a}_n(t)=a_{\text{mo}}(m)$, robot~$n$ is monitoring target~$m$,
which satisfies:
\begin{equation}\label{eq:monitor}
    d(\mathbf{y}_m(t), \mathbf{x}_n(t))\leq R_n, T_d(a_{\text{mo}}(m)) \geq T_0,
\end{equation}
where $d(\cdot)$ is the distance function, 
and~$T_0$ is the required minimum interval.
If $\mathbf{a}_n(t)=a_{\text{ch}}(s)$, robot $n$ charges at charging station $s$, 
i.e.,
\begin{equation*}
    b_n(t) \triangleq \begin{cases}
        \textbf{min}\left\{b_n^\text{max}, b_n(t-1)+\beta_s\right\},
     \hfill \textbf{if}\; \mathbf{a}_n(t)=a_{\text{ch}}(s);\\
     \textbf{max}\{b_n(t-1)-\gamma_n(t-1),0\}, \hfill \textbf{otherwise},
     \end{cases}
\end{equation*}
where the battery should satisfy the constraint:
\begin{equation} \label{eq: battery}
    0\leq b_n(t)\leq b_n^\text{max}, \forall{n} \in \mathcal{N}, \forall{t}\geq 0,
\end{equation}
which holds for all robots.
Moreover, let $\mathcal{T}_n^{\text{tar}} \triangleq \mathcal{M} \cap \mathcal{T}_n $.
There is a maximum number of targets any robot~$n$ can track simultaneously,
due to limited computation time, i.e., $|\mathcal{T}_n^{\text{tar}}|\leq C_n$.
The monitoring interval is fixed as $\chi_m$ for each target $m \in \mathcal{M}$.
The time from the last monitoring to the current time is denoted
by $\chi_m(t)$ that:
\begin{equation} \label{eq: t_m}
    0\leq \chi_m(t) \leq \chi_m, \forall{m} \in \mathcal{M}, \forall{t}\geq0,
  \end{equation}
  where $\chi_m(t)$ is set to zero if~$m$ is monitored, i.e.,
  if there exists~${n}\in\mathcal{N}_{\texttt{A}}(t)$,
  such that~$\mathbf{a}_n(t)=a_{\text{mo}}(m)$, $\chi_m(t)$ is set to zero.
  More importantly, each target should be monitored at intersections,
  i.e., if there exists~$t_0$, such that~$\chi_m^{\text{inter}}(t_0)=0, m \in\mathcal{M}$,
  then there exists~${n}\in\mathcal{N}_{\texttt{A}}(t),\enspace \mathbf{a}_n(t_0)=a_{\text{mo}}(m)$, where~$\chi_m^{\text{inter}}$ is the time of $m$ to the next intersection.
  Last but not least,
  the  rate of battery consumption~$\gamma_n(\cdot)$ is assumed to satisfy:
\begin{equation}    \label{eq: gamma_constraint}
    \gamma_n(t)=\gamma(||\mathbf{v}_n(t)||) \leq 
    \frac{b_n^{\text{max}}||\mathbf{v}_n(t)||}
    {2\cdot \sup_{\mathbf{w}\in \mathcal{W}} d_S(\mathbf{w})},
\end{equation}
where~$d_S(\mathbf{p})\triangleq \min_{s\in\mathcal{S}}d(\mathbf{p},\mathbf{z}_s)$.
This condition  ensures that any robot can visit
any position within~$\mathcal{W}$ and return to one charging station
with its maximum level of battery.

\subsection{Problem Statement} \label{subsec: statement}
The considered problem can be stated as a long-term constrained 
optimization problem, i.e.,
\begin{equation}\label{eq: problem}
\min_{\mathcal{N}_{\texttt{A}}(t),\{\mathbf{u}_n(t),\mathbf{a}_n(t)\}} 
    \lim_{T \to \infty} \frac{1}{T} \sum_{t=0}^{T} |\mathcal{N}_{\texttt{A}}(t)|,
\end{equation}
where the decision variables are the set of active robots and their local plans;
the objective is to minimize the average number of active robots;
and the constraints are~\eqref{eq:dynamics}-\eqref{eq: gamma_constraint}.

\section{Proposed Solution}\label{sec:solution}

The proposed solution is mainly composed of two layers: the search-based optimization for robot ordering, and the multi-robot maximum-allowed Martin's algorithm. 
In addition, the online execution, adaptations, and generalizations of the method are discussed. For convenience, the start time is set as $t=0$, and the ending time of the planning horizon is $\max_{m\in \mathcal{M}}T_m$.

\subsection{Search-based Optimization for Robot Ordering}
As an essential input to the following component,
the relative ordering among the robots is determined via a search-based scheme.
More specifically,
a search tree is constructed via iterative node selection and expansion.
Each node is a partial ordering of the robots
i.e.,~$\boldsymbol{\nu}=n_1n_2\cdots n_K$,
where~$\nu_k \in \mathcal{N}$ is a robot ID.
The cost of a node is defined as a vector:
$$\texttt{Cost}(\nu)\triangleq
\Big{(}|\mathcal{N}_{\texttt{A}}(t)|,
\underset{{n\in \mathcal{N}_{\texttt{A}}}}{\text{max}}\, T_{\boldsymbol{\tau}_n},
\sum_{n\in \mathcal{N}_{\texttt{A}}} \Delta b_{n}\Big{)},$$
consisting of the number of active robots,
the longest execution time of all active robots,
and the total battery consumption.
The root node is an empty sequence.
\textbf{Selection}: the node with the minimum makespan is selected
within the set of existing nodes.
\textbf{Expansion}: once a node is selected, it is expanded by
adding an additional robot to the existing sequence.
Consequently, the targets are assigned to the given sequence of
robots via the multi-robot maximum-allowed Martin's algorithm (MAM),
as described in the sequel.
Along with the sequence of targets, the node cost is also returned.
This node is called feasible if all targets are assigned.
Then, the procedure of selection and expansion is repeated,
until the planning time elapsed or all robot sequences are exhausted in the search tree.
To improve search efficiency, a branch-and-bound procedure
can be applied.
The lower bound on the makespan of each node
is computed as the minimum makespan when all robots are active,
while the upper bound is given by the one-step greedy assignment
where each target is assigned to the nearest robot if feasible.

\subsection{Multi-robot Maximum-allowed  Martin's Algorithm}
\begin{algorithm}[!t]
    \DontPrintSemicolon
    \KwIn{$\mathcal{R}, \mathcal{S},\mathcal{N},\mathcal{M}$, robot sequence $\nu$ }
    \KwOut{ $\mathcal{N}_{\texttt{A}}(0)$ and $\mathcal{T}_n,\forall{n}\in \mathcal{N}_{\texttt{A}}(0)$}
    \BlankLine
    Construct $G=(V,E)$ using $\mathcal{R}, \mathcal{S},\mathcal{N},\mathcal{M}$\label{line:graph}\;
    Initialize $\mathcal{N}_{\texttt{A}}(0)\gets \varnothing$\;
    Initialize targets-left set $V_{l}\gets V_{\text{tar}}$\label{line:target-left}\;
    \While{$V_l$ is not empty}{
        Robot $n\gets \alpha.\operatorname{pop}()$\;
        $V_{\text{tar}}^n, \mathcal{T}_n\gets$ Incremental-MA($V_l$, $n$)\label{line:IMA}\;
        \If{$\mathcal{T}_n\neq \varnothing$}{
        $V_l\gets V_l\setminus V_{\text{tar}}^n$\;
        $\mathcal{N}_{\texttt{A}}(0)\gets \mathcal{N}_{\texttt{A}}(0)\cup\{n\}$}
    }
    \textbf{Return} $\mathcal{N}_{\texttt{A}}(0),\mathcal{T}_n, \forall{n}\in \mathcal{N}_{\texttt{A}}(0)$
    \caption{Multi-robot MAM}
    \label{alg:overall}
\end{algorithm}
Given the robot sequence $\nu$, Alg.~\ref{alg:overall} is proposed to determine the active robot set $\mathcal{N}_{\texttt{A}}(0)$ and the optimal task sequence $\mathcal{T}_n$ for each active robot. The process begins by constructing the Time-Varying Node Network (TVNN) in Line~\ref{line:graph}, followed by executing Incremental Martin's Algorithm (IMA) for each robot in sequence in Line~\ref{line:IMA}, which outputs the maximum set of targets each robot can monitor along with the optimal $\mathcal{T}_n$. Finally, the minimum active robot set is obtained through a greedy approach.

\subsubsection{Time-Varying Node Network}
Robots $\mathcal{N}$, targets $\mathcal{M}$, and stations $\mathcal{S}$ are modeled as nodes in a temporal graph $G = (V, E)$, where $V$ is the node set and $E \subset V \times V$ is the edge set. The corresponding node sets are $V_N = \{\bar{n} \mid n \in \mathcal{N} \}$, $V_M = \{\bar{m} \mid m \in \mathcal{M} \}$, and $V_S = \{\bar{s} \mid s \in \mathcal{S} \}$, with $\bar{\cdot}$ distinguishing nodes from entities. To incorporate intersection constraints from \ref{subsec:constraint} and charging stations, special nodes are introduced or modified as follows:

\textbf{Intersection Target Nodes.} Intersection constraints introduce independent time constraints beyond~\eqref{eq: t_m}, requiring a virtual node $\bar{m}_i$ for each target $m \in \mathcal{M}$ to form $V_{M}^{\text{inter}}$. The target node set is thus $V_{\text{tar}} = V_M \cup V_{M}^{\text{inter}}$. Not all targets require both $\bar{m} \in V_M$ and $\bar{m}_i \in V_{M}^{\text{inter}}$. If $\chi_m^{\text{inter}} < \chi_m-\chi_m(0)$, $\bar{m} \in V_M$ is unnecessary as intersection monitoring already satisfies \cref{eq: t_m}. If $\chi_m^{\text{inter}}$ exceeds the planning horizon, $\bar{m}_i \in V_{M}^{\text{inter}}$ is not needed.
\textbf{Decomposition of Stations.}Upon reaching a charging station $s$, a robot selects its charging duration. To model this, each station node $\bar{s}$ is decomposed into a \textit{docking node} $\bar{s}_0$ and $N_s$ \textit{charging nodes} $\bar{s}_i$ for $i=1, \dots, N_s$. The notation $\bar{s} = {\bar{s}_i \mid i = 0, \dots, N_s}$ is used to denote an inseparable unit, with $V_S^{\text{dock}}$ and $V_S^{\text{charge}}$ representing all docking and charging nodes, respectively.
 The $i^{\text{th}}$ charging node supplies $b_n^{\text{max}} \cdot i/N_s$, providing $N_s$ charging options and enabling automated charge planning.

Let $V = V_N \cup V_{\text{tar}} \cup V_S$, where each node $\bar{v} \in V$ has a time-varying position: $\mathbf{p}_{\bar{n}}(t) = \mathbf{x}_n(0)$ for $\bar{n} \in V_N$, $\mathbf{p}_{\bar{m}}(t) = \mathbf{y}_m(t)$ for $\bar{m} \in V_M$, $\mathbf{p}_{\bar{m}_i}(t) = \mathbf{y}_m^{\text{inter}}$ for $\bar{m}_i \in V_M^{\text{inter}}$, where $\mathbf{y}_m^{\text{inter}}$ is the position of the next intersection of $m$, and $\mathbf{p}_{\bar{s}}(t) = \mathbf{z}_s$ for $\bar{s} \in V_S$. The resource of interest is $\mathbf{R} \in \mathbb{R}^2$, where $R^{(1)}$ is time and $R^{(2)}$ is battery consumption. Each node $\bar{v}$ has time and battery constraints $[R_{v,\text{min}}^{n,(i)}, R_{v,\text{max}}^{n,(i)}]$ for $i=1,2$, depending on robot $n$. Specifically, for $\bar{v} \in V_N \cup V_S$, the time constraint is $[0, +\infty]$ and the battery constraint is $[0, b_n^{\text{max}}]$. For $\bar{m} \in V_M$, the time constraint is $[0, \chi_m - \chi_m(0)]$, and the battery constraint is $[0, b_n^{\text{max}} - \gamma_n d_S(\mathbf{p}_{\bar{m}}(t))/v_n]$. For $\bar{m}_i \in V_M^{\text{inter}}$, the time constraint is $[\chi_m^{\text{inter}}(0)-T_0, \chi_m^{\text{inter}}(0)]$, and the battery constraint is $[0, b_n^{\text{max}} - \gamma_n d_S(\mathbf{p}_{\bar{m}_i}(t))/v_n]$.

Lastly, for connectivity, we define the set of reachable nodes $E_v \subset V$ for each node $v \in V$, i.e., $E_v = \{u \in V : (v, u) \in E\}$. Each edge $(v, u) \in E$ has an associated cost vector $\mathbf{C}(v, u) \in \mathbb{R}^2$. The connectivity and edge costs are summarized as follows.
\begin{equation} \label{eq: cost_connec}
\begin{aligned}
&C^{(1)}(\bar{v},\bar{u})=d(\mathbf{p}_{\bar{v}}(t),\mathbf{p}_{\bar{u}}(t))/v_n + T_0\cdot\mathbb{I}(\bar{v}\in V_{\text{tar}}), \\
    &C^{(2)}(\bar{v},\bar{u})=\gamma_n C^{(1)}(\bar{v},\bar{u}), \forall \bar{v}\in V\setminus V_S^{\text{dock}}, \bar{u}\in E_{\bar{v}},\\
    &C^{(1)}(\bar{s}_0,\bar{s}_i)=-C^{(2)}(\bar{s}_0,\bar{s}_i)/\beta_s, \\
  &C^{(2)}(\bar{s}_0,\bar{s}_i)=-b_n^{\text{max}}\cdot i/N_s,\forall \bar{s}_0\in V_S^{\text{dock}}, \bar{s}_i\in E_{\bar{s}},
\end{aligned}
\end{equation}
where~$E_n = V \backslash V_N, \, \forall n \in V_N,  E_{\bar{v}} = V \backslash V_N \cup \{\bar{v}\}, \, \bar{v} \in V_{\text{tar}}$;
$E_{\bar{s}_0} = \bar{s} \backslash \bar{s}_0, \, \forall \bar{s}_0 \in V_S^{\text{dock}},  E_{\bar{s}_i} = V \backslash V_N \cup V_S^{\text{dock}}, \, \forall \bar{s}_i \in V_S^{\text{charge}}$;
and the $\mathbb{I}(\cdot)$ is the indicator function.

\subsubsection{Incremental Martin's Algorithm}
\begin{algorithm}[!t]
    \DontPrintSemicolon
    \KwIn{Target left set $V_l$, robot $n$ }
    \KwOut{$V_{\text{tar}}^n, \mathcal{T}_n$}
    \BlankLine
    Initialize $V_N^n\gets \{\bar{n}\},V_{\text{tar}}^n\gets \varnothing, V_S^n\gets \varnothing$\;
        Initialize $L_{\bar{n},p}\gets\{\},L_{\bar{n},t}\gets\{\}$ \label{line: initialize1}\;
        $l_0 \gets (\bar{n}, \mathbf{R}_{l_0}, \varnothing)$\;
        $L_{\bar{n},t}\gets L_{\bar{n},t}\cup\{l_0\}$\label{line: initialize2}\;
        $\mathcal{T}_n\gets \varnothing$\;
         \While{$V_l$ is not empty}{
            $\bar{c}\gets \arg\min_{\bar{v}\in V_l}\kappa(\bar{v}, \bar{n})$\label{line: new-target}\;
            $\bar{s}\gets \arg\min_{\bar{s}\in V_S} d(\mathbf{z}_s, \mathbf{p}_{\bar{c}}(0))$\label{line: new-station}\;
            $V_{\text{tar}}^n\gets V_{\text{tar}}^n\cup\{\bar{c}\}$, $V_S^n\gets V_S^n\cup\{\bar{s}\}$\;
            Get $E^n$ according to~\eqref{eq: cost_connec}\;
            $L_{\bar{c},p}\gets\{\},L_{\bar{c},t}\gets\{\},L_{\bar{s},p}\gets\{\},L_{\bar{s},t}\gets\{\}$\label{line:new-labelset}\;
            Add a new dimension of 1 to $\mathbf{\hat{R}}_{l_{\bar{v}}}, \forall l_{\bar{v}}\in L_p^n\cup L_t^n$\label{line: new-dimension}\;
            Propagate all nodes in $L_p^n$ to $\bar{c}$ and $\bar{s}$\label{line:supple-prop}\;
            $\mathcal{A}\gets\operatorname{MP}(G^n, L_p^n,L_t^n)$\label{line:MA}\;
            \If{$\mathcal{A}=\varnothing$}{
            $V_{\text{tar}}^n\gets V_{\text{tar}}^n\setminus\{\bar{c}\}$, $V_S^n\gets V_S^n\setminus\{\bar{s}\}$\;
                \textbf{Break}\;
            }
            $\mathcal{T}_n\gets \mathcal{A}$\;
         }
    \textbf{Return} $V_{\text{tar}}^n, \mathcal{T}_n$\;
    \caption{Incremental Martin's Algorithm}
    \label{alg:IMA}
    \end{algorithm}
Given the target left set $V_l$, robot $n$ constructs an incremental sub-TVNN $G^n = (V^n, E^n)$, where $V^n = V_N^n \cup V_{\text{tar}}^n \cup V_S^n$ with $V_{\text{tar}}^n \subset V_{\text{tar}}$ and $V_S^n \subset V_S$. In each iteration, the algorithm selects the optimal target node and its nearest charging station, adding them to $V^n$. Target priority is determined by $\kappa(\cdot, \bar{n})$ in Line~\ref{line: new-target}, i.e.,
\begin{equation}
    \kappa(\bar{v}, \bar{n}) = a_1\cdot d(\mathbf{p}_{\bar{v}}(0), \mathbf{p}_{\bar{n}}(0))/v_n+a_2\cdot R_{\bar{v},\text{max}}^{n,(1)},
\end{equation}
where $\bar{v}\in V_{\text{tar}}^n$ and $a_1.a_2\in \mathbb{R}^+$ are coefficients. The function $\kappa(\cdot,\bar{n})$ is defined by the travel time from $\bar{n}$ to $\bar{v}$ and the time constraint of $\bar{v}$, ensuring that nodes closer to $\bar{n}$ and with more urgent time constraints are prioritized.

The IMA follows a \emph{label-setting} approach, utilizing labels to explore the graph and identify the Pareto front. Each label $l$ is represented as a tuple $(\bar{v},\mathbf{R}_{l},\bar{c})$, where $\bar{v}$ is the node, $\mathbf{R}_l$ is the resource vector, and $\bar{c}$ is the predecessor node. Given the multidimensional nature of $\mathbf{R}_l$, the \textit{dominance} rule is applied for comparison. Specifically, ``$\mathbf{a}$ dominates $\mathbf{b}$'' is denoted as ``$\mathbf{a} \prec_{\text{P}} \mathbf{b}$'', meaning $\mathbf{b}$ does not yield a better outcome than $\mathbf{a}$. This relation holds if $(a_1, \ldots, a_n)^T \prec_{\text{P}} (b_1, \ldots, b_n)^T \Leftrightarrow \mathbf{a} \neq \mathbf{b} \land a_i \leq b_i, \forall i \in \{1, \ldots, n\}$. Labels are managed in permanent sets $L_{v,p}$ and temporary sets $L_{v,t}$ for each node $v\in V$. The set $L_{v,p}$ contains Pareto-optimal labels at $v$, while $L_{v,t}$ holds intermediate labels. For convenience, we define the total permanent and temporary sets as $L_p^n=\bigcup_{\bar{v}\in V^n}{L_{\bar{v},p}}$ and $L_t^n=\bigcup_{\bar{v}\in V^n}{L_{\bar{v},t}}$, respectively.

To ensure all target nodes are visited in each plan, we introduce a binary vector $\mathbf{\hat{R}}\in\mathbb{R}^{N_{\text{tar}}}$, where $N_{\text{tar}}=|V_{\text{tar}}^n|$, to indicate the visitation status of each node in $V_{\text{tar}}^n$. Each target node $\bar{v} \in V_{\text{tar}}^n$ is assigned an index $i_{\bar{v}} \in \{1,2,\dots, N_{\text{tar}}\}$, with each dimension of $\mathbf{\hat{R}}$ corresponding to a target node. A value of $1$ denotes an unvisited node, while $0$ indicates it has been visited. The initial resource state is set as $\mathbf{\hat{R}}_{l_0}^{\bar{n}}=\mathbf{1}_{N_\text{tar}}$. The cost vectors associated with node resources are given by

\begin{equation}
    \mathbf{\hat{C}}(\bar{v}, \bar{u}) = -\mathbf{e}_{i_{\bar{u}}}\mathbb{I}(\bar{u}\in V_{\text{tar}}^n),
\end{equation}
where $\mathbf{e}_i\in \mathbb{R}^{N_{\text{tar}}}$ is a unit vector with 1 in the $i^{\text{th}}$ dimension and 0 elsewhere. Since revisiting target nodes is unnecessary, the resource constraint $\hat{R}^{(i)} \geq 0$ is imposed to expedite the search process. By concatenating the respective resources and costs, the final resource and cost are obtained as:

\begin{equation}
\mathbf{\tilde{R}}=
    \begin{pmatrix}
        \mathbf{R} \\
        \mathbf{\hat{R}}
    \end{pmatrix},
\mathbf{\tilde{C}}=
    \begin{pmatrix}
        \mathbf{C} \\
        \mathbf{\hat{C}}
    \end{pmatrix}.
\end{equation}


After preparing new nodes in Lines \ref{line:new-labelset}-\ref{line: new-dimension}, all permanent nodes are propagated to $\bar{c}$ and $\bar{s}$, and Martin's Process (MP), inspired by Martin's Algorithm \cite{martins1984multicriteria}, is executed to determine the optimal task sequence $\mathcal{A}$ within the current sub-network $G^n$. If MP fails to return a feasible solution, the maximal set of targets monitored by robot $n$ is obtained by removing the newly added nodes, reverting to the optimal $\mathcal{T}_n$ from the previous iteration.

\subsubsection{Martin's Process}\label{subsubsec:MA}
\begin{algorithm}[!t]
    \DontPrintSemicolon
    \KwIn{$G^n=(V^n,E^n),L_p^n,L_t^n$}
    \KwOut{Multiobjective shortest path $\mathcal{A}$}
    \BlankLine
    \While{$L_t^n$ is not empty}{ \label{line:while}
        $l_{{\bar{c}}}\gets\arg\min_{l_{\bar{v}}\in L_t^n}^{\prec_{\text{lex}}} \mathbf{R}_{l_{\bar{v}}}$
        \label{line:pop}\;
        $\bar{c}\gets l_{\bar{c}}.\operatorname{Node}()$\;
        \If{$\operatorname{terminate}(l_{\bar{c}})$}{ \label{line:terminate}
            \textbf{Return} $\mathcal{A}\gets \operatorname{backtracking}(l_{\bar{c}})$
            \label{line:backtracking1}\;
        }
        $L_{{\bar{c}},t}\gets L_{{\bar{c}},t}\setminus \{l_{\bar{c}}\}$ \label{line:remove}\;
        $L_{{\bar{c}},p}\gets L_{{\bar{c}},p}\cup \{l_{\bar{c}}\}$ \label{line:insert}\;
        \For{$\bar{v} \in E_{\bar{c}}$}{
        \tcp{Propagate to $\bar{v}$}
            $\mathbf{\tilde{R}}_{l_{\bar{v}}}\gets \mathbf{\tilde{R}}_{l_{\bar{c}}}+\mathbf{\tilde{C}}(\bar{c},\bar{v})$\label{line: propagates}\;
            \If{$R_{l_{\bar{v}}}^{(i)} \leq R_{\bar{v},\text{max}}^{n,(i)},\forall{i}\leq 2$ and $\hat{R}_{l_{\bar{v}}}^{(i)}\geq 0,\forall i\leq N_{\text{tar}}$\label{line: resource1}}{
                $R_{l_{\bar{v}}}^{(i)}\gets \max\{R_{l_{\bar{v}}}^{(i)},R_{\bar{v},\text{min}}^{n,(i)}\},
                \forall{i}\leq 2$ \label{line: resource2}\;
            $l_{\bar{v}}\gets (\bar{v}, \mathbf{\tilde{R}}_{l_{\bar{v}}},\bar{c})$\;
            \If{$\nexists \, l\in L_{\bar{v},p} \cup L_{\bar{v},t} \text{ s.t. } l \prec_{\text{P}} l_{\bar{v}}$\label{line: dominance1}}{
                $L_{\bar{v},t}\gets L_{\bar{v},t} \setminus \{l\in L_{\bar{v},t}: l_{\bar{v}}\prec_{\text{P}} l\}$\label{line: dominance2}\;
                $L_{\bar{v},t} \gets L_{\bar{v},t} \cup \{l_{\bar{v}}\}$ \label{line: new-insert}\;
            }
            }
        }

    }
    {\textbf{Return} $\varnothing$}
    \caption{Martin's Process (MP)}
        \label{alg:MA}
    \end{algorithm}
In each iteration, the optimal $l_c$ is selected from $L_t^n$, transferred into $L_p^n$, and propagated from $\bar{c}$ to all its viable nodes.
The optimization objective is defined by the lexicographic order comparison function $\prec_{\text{lex}}$ (Line~\ref{line:pop}).
This multidimensional comparison rule prioritizes lower time consumption while also minimizing battery usage. Notably, $\prec_{\text{lex}}$ satisfies the properties of \textit{dominance} and \textit{monotonicity}, ensuring that labels are extracted in the correct order \cite{paixao2013labeling}. The operation $\min^{\prec_\text{lex}}$ in Line \ref{line:pop} refers to the minimum selection based on the $\prec_{\text{lex}}$ ordering.
The \emph{propagate} process begins by generating a new resource vector through cost addition in Line~\ref{line: propagates}. If the new vector is valid (Line~\ref{line: resource1}), it is  adjusted by $R_{v,\text{min}}^{n,(i)}$ in Line~\ref{line: resource2} and create a new label. The right boundary of the resource window is a strict constraint that cannot be exceeded, while the left boundary, if not reached, is compensated, representing waiting time for $i=1$ and invalid excess charge for $i=2$. If the new label is non-dominated at $\bar{v}$, dominated labels in the temporary set are removed, and the new label is inserted. When the optimal label $l_{\bar{c}}$ meets the early termination condition, i.e., $\mathbf{\hat{R}}_{l_{\bar{c}}}=\mathbf{0}_{N_{\text{tar}}}$, the optimal task sequence $\mathcal{A}$ is obtained by backtracking each label’s predecessor from $l_{\bar{c}}$.



\subsection{Online Execution}\label{sec:online}
\subsubsection{Online Execution and Adaptations}
We define a Replanning Horizon $T_h$
to ensure timely corrective actions when constraints are at risk of being violated. By analytically computing the exact time at which each target is expected to violate its constraints, constraint satisfaction can be proactively monitored. If any violation is predicted to occur within $T_h$
, a replanning process is triggered to maintain feasibility. The duration of $T_h$
is chosen such that the UAV can reach any drone on the field, ensuring that necessary interventions can always be executed in time.

We modify Alg.~\ref{alg:overall} and Alg.~\ref{alg:IMA} in two aspects. First, in Line~\ref{line:target-left}, instead of considering the entire set $V_{\text{tar}}$, only targets that violated constraints in prediction, have completed monitoring, or were newly generated are included. Second, when constructing the sub-TVNN, ongoing targets from the previous allocation are prioritized and incorporated first, followed by the remaining targets in $V_l$. These modifications ensure recursive feasibility, as targets in $V_l$ must satisfy $R_{v, \text{max}}^{n, (i)} \geq T_h$ due to the triggering condition.
During online execution, at each time step
    $t\geq0$, the UAV follows the current plan and determines its actions accordingly. Then the robots will compute the corresponding $\mathbf{u}_n(t)$ and move according to~\eqref{eq:dynamics}, which can be computed through the \emph{NMPC} in \cite{wang2024uncertainty}, and the targets move according to their kinematics and choose the next road randomly at the meantime. This process continues until a target is predicted to violate its constraints, at which point the next replanning event is triggered to ensure feasibility.

\subsubsection{Complexity Analysis}
    According to \cite{de2021improved}, the complexity of IMA is $\mathcal{O}((N_{\text{tar}}+N_s |V_S^n|) N_{\text{tar}} N_{\text{label}}^2)$, where $N_{\text{label}}$ denotes the number of non-dominated labels at the algorithm's termination.
Since $N_{\text{tar}}$ and $|V_S^n|$ are bounded by $C_n$, the complexity simplifies to $\mathcal{O}(N_s C_n^2 N_{\text{label}}^2)$.
Thus, the complexity of MAM is $\mathcal{O}(|\mathcal{N}_{\texttt{A}}| N_s C_n^2 N_{\text{label}}^2)$.

\subsection{Generalization}
The proposed planning framework can be generalized in the following aspects:
(I) \emph{Robot Failures}.
In the event of a robot failure, e.g.,
an unexpected UAV crash,
this event is immediately treated as a triggering condition for online adaptation.
Consequently,
all targets assigned to the crashed UAV are added to $V_l$ as the set of unassigned targets.
A replanning is initiated to redistribute these tasks,
via the same adaptation algorithm in Sec.~\ref{sec:online}.
(II) \emph{Free Membership of Targets}
In case new targets enter the workspace or existing targets leave the workspace,
the proposed framework can still be applied.
Namely, any target that leaves the workspace during execution is removed from the Node Network in real time,
which automatically is not considered in the assignment.
Similarly, when a new target enters the scene,
it is detected by at least one robot and added to the set of targets to be assigned.
Then, the same procedure can be applied.


\section{Numerical Experiments} \label{sec:experiments}
For further validation,
extensive numerical simulations are conducted.
The algorithm is implemented in \texttt{Python3}
and tested on a laptop with an Intel Core i7-1280P CPU.
Simulation videos can be found in the supplementary files.

\subsection{System Setup}
\begin{figure}[!t]
    \centering
   \includegraphics[width = \linewidth]{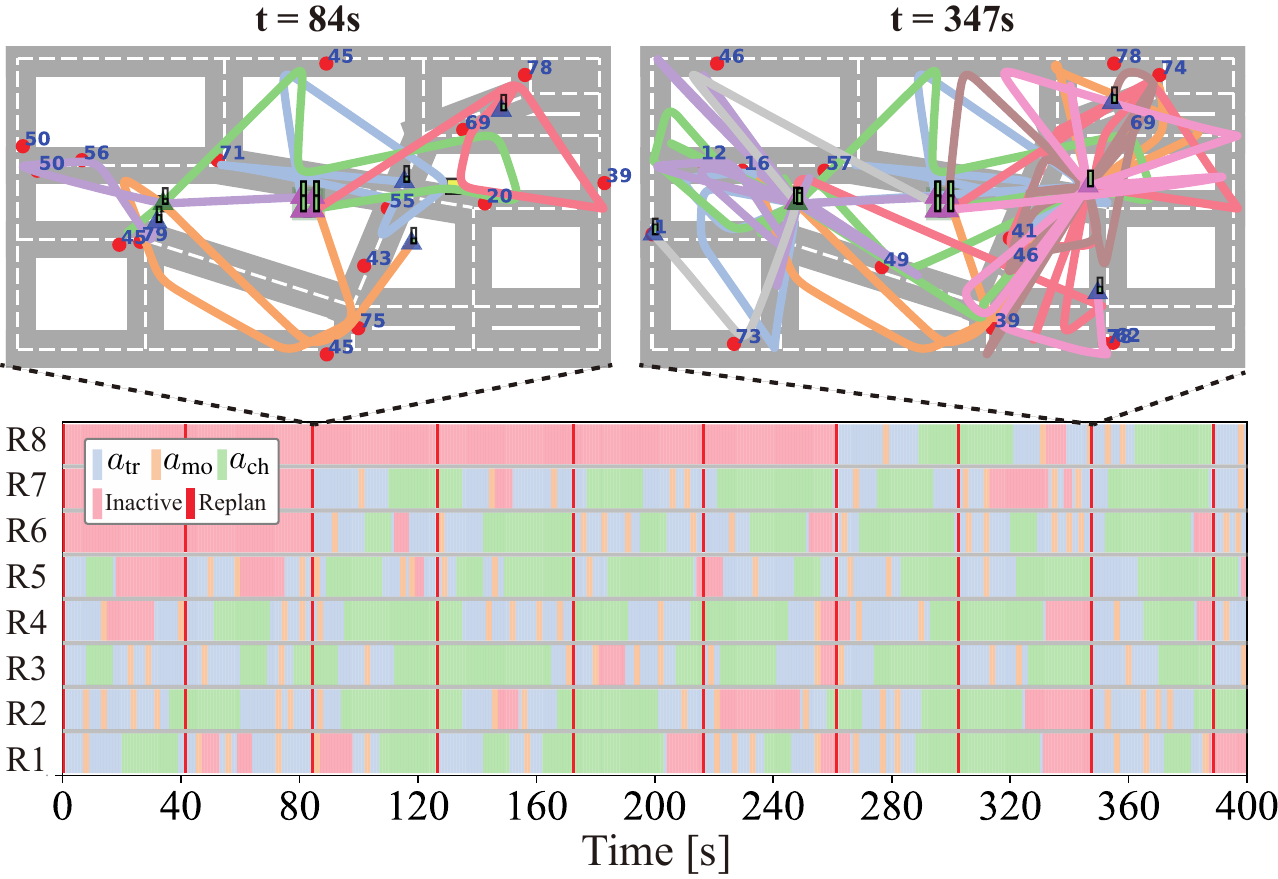}
    \vspace{-0.7cm}
    \caption{\textbf{Top-Left} and \textbf{Top-Right}: Trajectories of 8 robots actively monitoring 15 targets in the considered scenario at t=$\SI{84}{s}$ and t=$\SI{347}{s}$ respectively.
    \textbf{Bottom}: The action of 8 robots at each time during the simulation of $\SI{400}{s}$.
    }
 	\label{fig:description}
     \vspace{-0.3cm}
\end{figure}
As shown in Fig.~\ref{fig:overall} and Fig.~\ref{fig:description}, our algorithm performs well in both scenarios.
In Fig.~\ref{fig:overall}, 10 targets move within a $\SI{50}{m} \times \SI{50}{m}$ road network with three charging stations. Each robot follows a linear UAV motion model at a fixed height, starting from the central drone pad with a fully charged battery, $b_n^{\text{max}} = 10.0$. There is no limit on the number of robots at the drone pad.
Robots have a maximum speed of $v_n^{\text{max}} = \SI{1.5}{m/s}$, a battery consumption rate of $\gamma_n = \SI{0.2}{s^{-1}}$, and a charging rate of $\beta_s = \SI{0.2}{s^{-1}}$. The monitoring range in (\ref{eq:monitor}) is $R_n = \SI{3}{m}$, and the capacity is uniformly $C_n = 6$ for all $n \in \mathcal{N}$.
Targets are initially distributed randomly along roads and choose their next segments randomly at intersections, unknown to the robots. Each target moves at $||\mathbf{v}_m|| = \SI{0.2}{m/s}$, with a maximum monitoring interval of $\chi_m = \SI{80}{s}$ and a monitoring duration of $T_0 = \SI{2}{s}$.
The number of charging nodes is $N_s = 5$, and the triggering horizon is $T_h = \SI{30}{s}$. The simulation runs for $\SI{400}{s}$.
In Fig.~\ref{fig:description}, the parameters are modified as follows: the road network size is $\SI{100}{m} \times \SI{50}{m}$, with two charging stations. The speed is set to $\SI{3}{m/s}$, and the number of targets is 15.

\subsection{Results}\label{subsec:results}

The final results are shown in Fig.~\ref{fig:overall} and Fig.~\ref{fig:description}.
In Fig.~\ref{fig:overall}, each replan employs an average of 3.7 robots, with a computational cost of $\SI{0.063}{s}$. The average planning time per robot $n \in \mathcal{N}_{\texttt{A}}$ is $\SI{0.020}{s}$, with 22.6 nodes in each robot's planning network. The average number of non-dominated labels at the end of each planning process is 115.4.
Overall, the average replan interval is $\SI{46.13}{s}$, and the average number of employed robots is 3.5. Each robot $n$ charges for an average of $\SI{109.5}{s}$, accounting for 49.6\% of its active period, indicating a high degree of autonomous charging management.
According to Fig.~\ref{fig:overall}, in both scenarios, the batteries of all robots remain positive, and the monitoring intervals of all targets stay below $\chi_m$ throughout the simulation. In the bottom-right figure of Fig.~\ref{fig:overall}, at least one robot is present in the two rows of each target during every replan, ensuring that each target remains in the task sequence of at least one robot at all times.

In the scenario depicted in Fig.~\ref{fig:description}, each replan employs an average of 5.5 robots, with an average computational cost of $\SI{0.164}{s}$. For each replan, the average planning time per robot $n \in \mathcal{N}_{\texttt{A}}$ is $\SI{0.024}{s}$, with an average of 16.5 nodes in each robot's planning network and 111.7 non-dominated labels at the end of planning.
Overall, throughout the simulation, the average replan interval is $\SI{43.11}{s}$, and the average number of employed robots is 6.2. Each robot $n$ spends an average of $\SI{143.5}{s}$ charging, accounting for 47.7\% of its active period.

\subsection{Comparisons}\label{subsec:comparisons}
\begin{figure*}[t]
    \centering
    \setlength{\abovecaptionskip}{0.cm}
    \includegraphics[width=\linewidth]{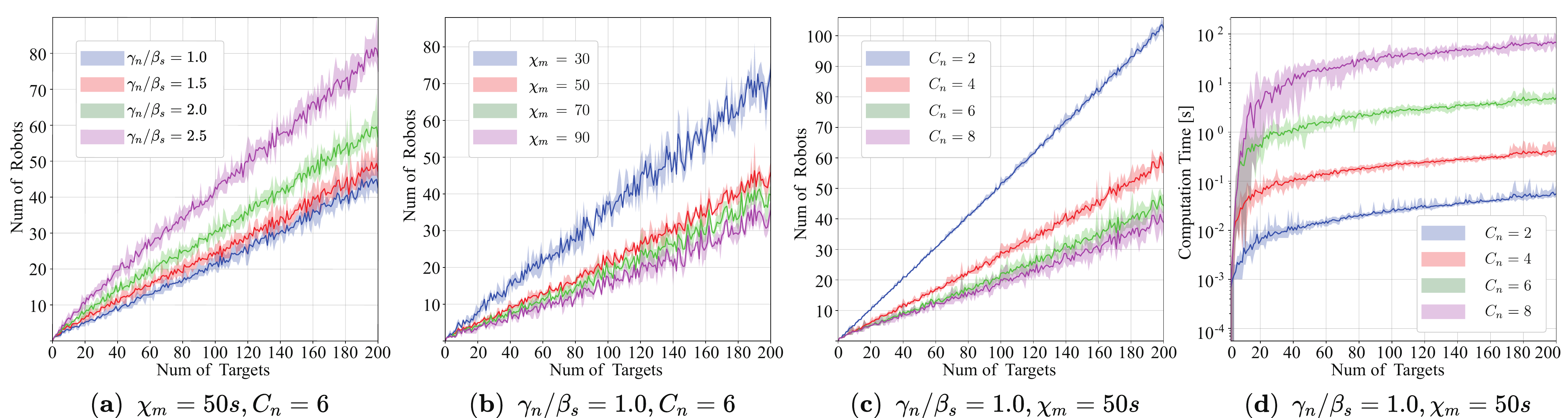}
    \vspace{-0.3cm}
    \caption{Illustration of scalability analysis results. The number of active robots with an increasing number of targets, concerning $\gamma_n/\beta_s$ \textbf{(a)}, $T_m$ \textbf{(b)} and $C_n$ \textbf{(c)}. The computation time with an increasing number of targets, concerning $C_n$ \textbf{(d)}.}
    \label{fig:scalability}
    \vspace{-0.3cm}
\end{figure*}
\begin{table}[t]
 \begin{center}
 \begin{threeparttable}
   \caption{Comparison analysis results}\label{table:comparison-data}
\label{table:comparison}
   \vspace{-0.05in}
   \setlength{\tabcolsep}{0.7\tabcolsep}
   \centering
   \renewcommand{\arraystretch}{1.1}
   \begin{tabular}{c c|c c c c}
     \toprule
     \makecell{\textbf{Cases}} & \textbf{Methods} & \makecell{\textbf{Succsess}\\ \textbf{Rate [\%]}} & \makecell{\textbf{Average}\\\textbf{$|\mathcal{N}_A|$}} & \makecell{\textbf{Num of}\\\textbf{Nodes}} & \makecell{\textbf{Computation}\\\textbf{Time [s]}}  \\
     \midrule
     \multirow{5}{*}{\makecell{$M$=8}} & Ours & 100 & 2.81  & 15.62 & 0.09 \\
      & NI-MAM & 100 & 2.54  & 24.30 & 77.98 \\
      & CMA  & 100 & 2.40 & 5228.70 & 2.60  \\
      & PeT & 86.7  & 2.90 & 15.36  & 0.10 \\
      & GCF & 63.4 & 3.38 & 2.34  & 0.01 \\
     \midrule
      \multirow{5}{*}{\makecell{$M$=30}} & Ours & 100 & 7.69  & 16.57 & 2.41 \\
      & NI-MAM & 100 & 7.34  & 32.59 & 470.57\\
      & CMA  & - *& $>6$ & 1.41e14 & $>600$ \\
      & PeT & 78.4 & 7.86  & 16.88 & 2.39 \\
      & GCF & 47.9 & 12.76  & 2.36 & 0.04\\
     \midrule
     \multirow{5}{*}{\makecell{$M$=100}} & Ours & 100 & 23.54  & 16.61 & 6.42  \\
      & NI-MAM & - & $>20$  & 118.51  & $>$600 \\
      & CMA  & - & $>20$  & 7.15e52  & $>$600 \\
      & PeT & 69.5 & 25.18  & 16.35  & 6.43 \\
      & GCF & 33.1 & 41.75  & 2.40  & 0.09 \\
     \bottomrule
      \multicolumn{6}{l}{\multirow{2}{*}{*``-'' means the result cannot be obtained within the limited time.} }
   \end{tabular}
 \end{threeparttable}
\end{center}
    \vspace{-0.1in}
 \end{table}
To validate the proposed framework (\textbf{Ours}), we conduct a quantitative comparison against four baselines, including two ablation studies and two common approaches:
(i)Non-incremental Maximum-allowed Martin's Algorithm (\textbf{NI-MAM}): Targets are not added incrementally for each robot but are all included in the sub-TVNN at each step. The robot returns a sequence maximizing visited targets.
(ii) Centralized Martin's Algorithm (\textbf{CMA}): All robots are treated as a single system, where Martin's algorithm is applied in a centralized manner~\cite{schillinger2017multi}.
(iii) Periodic Triggering (\textbf{PeT}): The triggering condition is replaced by periodic triggering.
(iv) Greedy Closest-First (\textbf{GCF}): Targets are assigned greedily, iteratively allocating each robot to the nearest available target until no further feasible assignments remain.


As summarized in Table~\ref{table:comparison}, the the success rate, average number of robots, average nodes per replan, and computation time are compared in the nominal setup of Fig.~\ref{fig:overall} with $C_n = 5$ and varying $M$.
Although our method uses slightly more robots than (i) and (ii), the average number of nodes remains stable (always $< 20$), and the computation time increases more gradually ($\SI{6.42}{s}$ at $M = 100$). In contrast, (i) and (ii) show a sharp increase in both metrics as $M$ grows, demonstrating higher scalability for our method.
The PeT method exhibits similar $\mathcal{N}_{\texttt{A}}$, average nodes, and computation time as ours but has a lower success rate (below 100\%) due to the absence of active replan time prediction. The GCF method has smaller average nodes and computation time due to its direct selection principle, but its success rate is lower and requires more robots.

\subsection{Scalability Analysis}

The scalability of the proposed algorithm is analyzed with respect to four aspects: the number of targets $M$, the ratio between battery consumption rate $\gamma_n$ and charging rate $\beta_s$, the upper bound of the interval $\chi_m$, and the capacity $C_n$. The results are summarized in Fig.~\ref{fig:scalability}, with the nominal setup shown in Fig.~\ref{fig:overall}.
From (a), (b), and (c), we observe that the ratio between the total number of targets and the average number of active robots remains nearly constant, which we define as \emph{fleet efficiency}, denoted as $\eta$, and is upper-bounded by $C_n$. In (a), we see that as the ratio of battery consumption to charging rate increases, fleet efficiency decreases from 4.5 to 2.5, as each robot can move less and track fewer targets under higher consumption-charging ratios. In (b), when the maximum interval $\chi_m$ increases from $\SI{30}{s}$ to $\SI{90}{s}$, fleet efficiency increases from 2.7 to 6.0, as a larger $\chi_m$ allows each robot to track more targets due to the extended tracking time.
From (c), we observe that as $C_n$ increases from 2 to 8, fleet efficiency rises from 1.9 to 5.0, as more targets can be monitored by each robot.
Figure (d) examines computation time for different $C_n$ values as the number of targets increases. Our method demonstrates strong scalability, as it takes less than $\SI{10}{s}$ to compute when $M = 200$ and $C_n = 6$. It also highlights that $C_n$ is a crucial factor influencing computation time, with an exponential relationship between computation time and $C_n$.

\subsection{Generalization}
\begin{figure}[t]
    \centering
    \includegraphics[width=\linewidth]{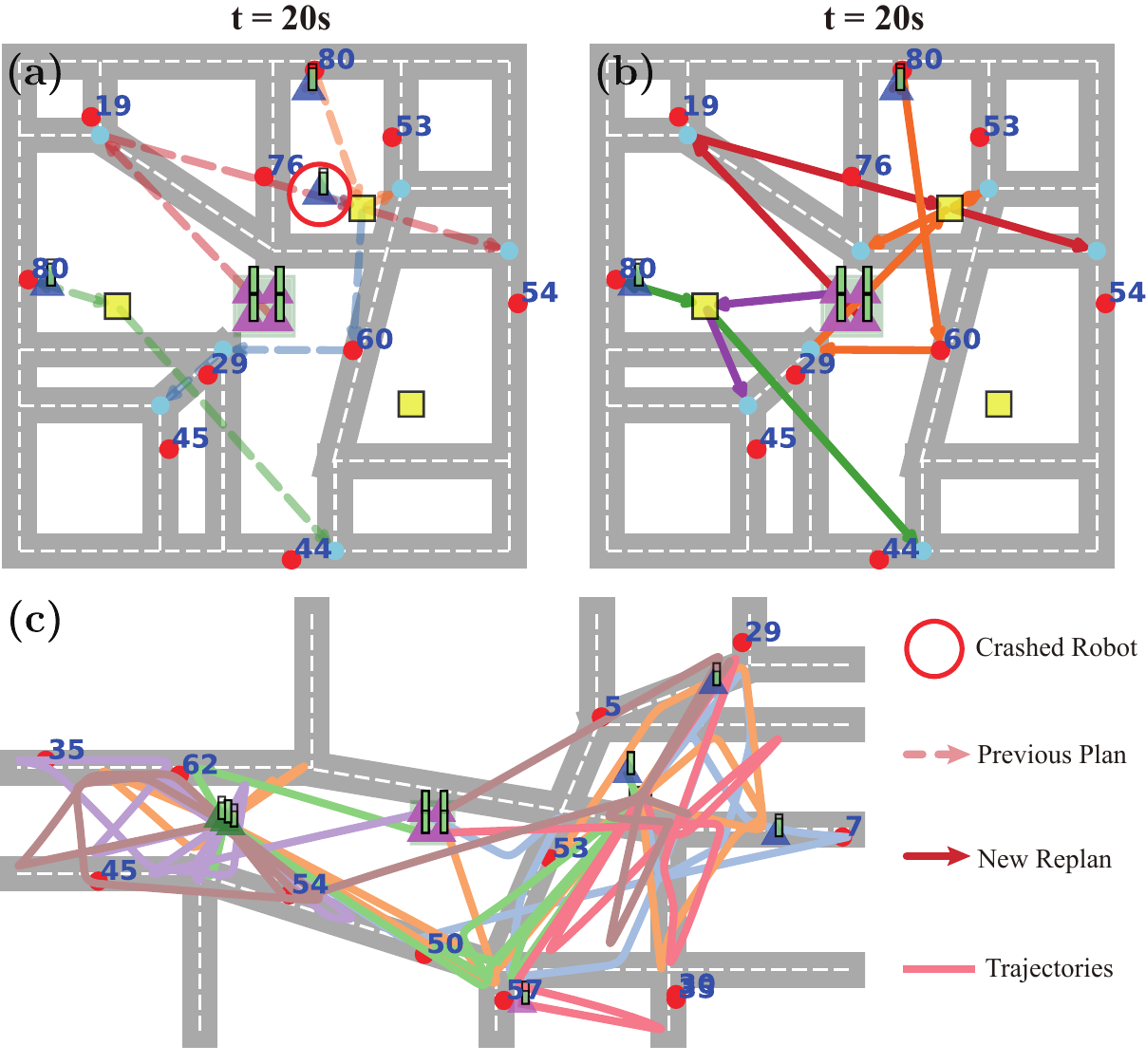}
    \vspace{-0.8cm}
    \caption{Illustration of the generalization circumstances. \textbf{(a)}: Previous plan result at $t=\SI{20}{s}$ when a robot accidentally crashes. \textbf{(b)}: New replan after the robot crashed. \textbf{(c)}: Trajectories of 6 robots monitoring 15 targets that can dynamically enter and exit the road network over a period of $\SI{400}{s}$}
    \label{fig:generalization}
    \vspace{-0.3cm}
\end{figure}
Performances under generalized scenarios are also validated as shown in Fig.~\ref{fig:generalization}.
Namely, one robot failed at~$t=20s$, after which
the targets it was tracking are reassigned to other robots in the new replan.
The adaptation takes~$0.02s$ and the local plans of the other robots are modified accordingly.
Moreover, $6$ targets enter freely into the workspace during online execution,
with the setup otherwise identical to Fig.~\ref{fig:description}.
The trajectories of the employed robots are adapted
to monitor the new targets.


\section{Conclusion} \label{sec:conclusion}
This work addresses the long-term monitoring of dynamic targets in a road network using a fleet of aerial robots with limited resources. We propose a hierarchical approach that incrementally assigns targets, optimizes monitoring sequences and charging strategies, and adapts online to real-time constraints. Our method ensures strict adherence to resource and monitoring constraints while minimizing the active fleet size. Extensive simulations demonstrate its scalability and effectiveness in deploying a small UAV fleet for large-scale target monitoring.


\bibliographystyle{IEEEtran}
\bibliography{references}

\end{document}